\documentclass[letterpaper]{article} 
\usepackage[preprint]{aaai2027}  
\usepackage[hyphens]{url}  
\usepackage{graphicx} 
\usepackage{natbib}  
\usepackage{caption} 
\usepackage{algorithm}
\usepackage{algorithmic}
\usepackage{amsmath}
\usepackage{amssymb}
\usepackage{newfloat}
\usepackage{listings}
\DeclareCaptionStyle{ruled}{labelfont=normalfont,labelsep=colon,strut=off} 
\floatstyle{ruled}
\newfloat{listing}{tb}{lst}{}
\floatname{listing}{Listing}

\usepackage{booktabs}

\title{HyperANFIS: Enhancing Rule Representation and Interpretability in Adaptive Neuro-Fuzzy Systems via Hyperbolic Geometry}
\author{
    Haoran Pei\textsuperscript{\rm 1},
    Zhao Su\textsuperscript{\rm 1},
    Zetao Lin\textsuperscript{\rm 1},
    Haoran Li\textsuperscript{\rm 2},
    Jun Shen\textsuperscript{\rm 3},
    Qi Zhu\textsuperscript{\rm 4},
    Lan Guo\textsuperscript{\rm 1},\\
    Qingguo Zhou\textsuperscript{\rm 1},
    Binbin Yong\textsuperscript{\rm 1}\thanks{Corresponding author: Binbin Yong(yongbb@lzu.edu.cn)}
}
\affiliations{
    \textsuperscript{\rm 1}school of Information Science and Engineering, Lanzhou University\\
    \textsuperscript{\rm 2}Department of Data Science and AI, Monash University\\
    \textsuperscript{\rm 3}School of Computing and Information Technology, University of Wollongong\\
    \textsuperscript{\rm 4}College of Artificial Intelligence, Nanjing University of Aeronautics and Astronautics\\

}

\begin{document}

\maketitle

\begin{abstract}
The adaptive neuro-fuzzy inference system (ANFIS) is an interpretable reasoning framework capable of generating explicit IF-THEN fuzzy rules, making it suitable for tasks requiring transparent reasoning. However, existing ANFIS models generally construct rule antecedents and perform inference in Euclidean space, limiting their representational capacity and predictive performance. To address this issue, we propose Hyperbolic ANFIS (HyperANFIS), a hyperbolic extension of ANFIS. HyperANFIS preserves the fuzzy semantics and core architecture of conventional ANFIS while performing rule-prototype learning, rule activation, and consequent aggregation in hyperbolic space. It also retains the ability to generate interpretable IF-THEN rules. By exploiting the representational properties of hyperbolic geometry, HyperANFIS strengthens the fuzzy inference process, thereby improving predictive accuracy, inter-rule collaboration, and the credibility of its interpretable rules. Experimental results show that HyperANFIS consistently outperforms the standard ANFIS baseline and various ANFIS variants across all datasets, while also generating higher-quality fuzzy rules.
\end{abstract}

\section{Introduction}

Adaptive neuro-fuzzy inference systems (ANFIS) are data-adaptive machine-learning models that combine the parameter-learning capability of neural networks with the rule-based inference of fuzzy systems. ANFIS models nonlinear input-output relationships through fuzzy IF-THEN rules parameterized by membership functions \cite{256541}. Because its rule base and rule activations can be inspected, ANFIS is particularly appealing in applications such as healthcare, industry, and finance, where transparent decision-making is important.

Despite these advantages, conventional ANFIS architectures face structural limitations as the number of input variables increases or as the data exhibit complex relational structures \cite{jin2024simplification}. In a classical ANFIS, the antecedent of each rule is typically formed by conjoining univariate membership functions defined on individual features in Euclidean space, with these membership functions treated independently, while the rule firing strength is obtained by aggregating their membership values. As the numbers of input variables and membership functions increase, the number of rules grows combinatorially. Meanwhile, the increasingly numerous feature representations become crowded and constrained in Euclidean space, which can affect model performance and rule-generation capacity.

\begin{figure}[t]
    \centering
    \includegraphics[width=1\linewidth]{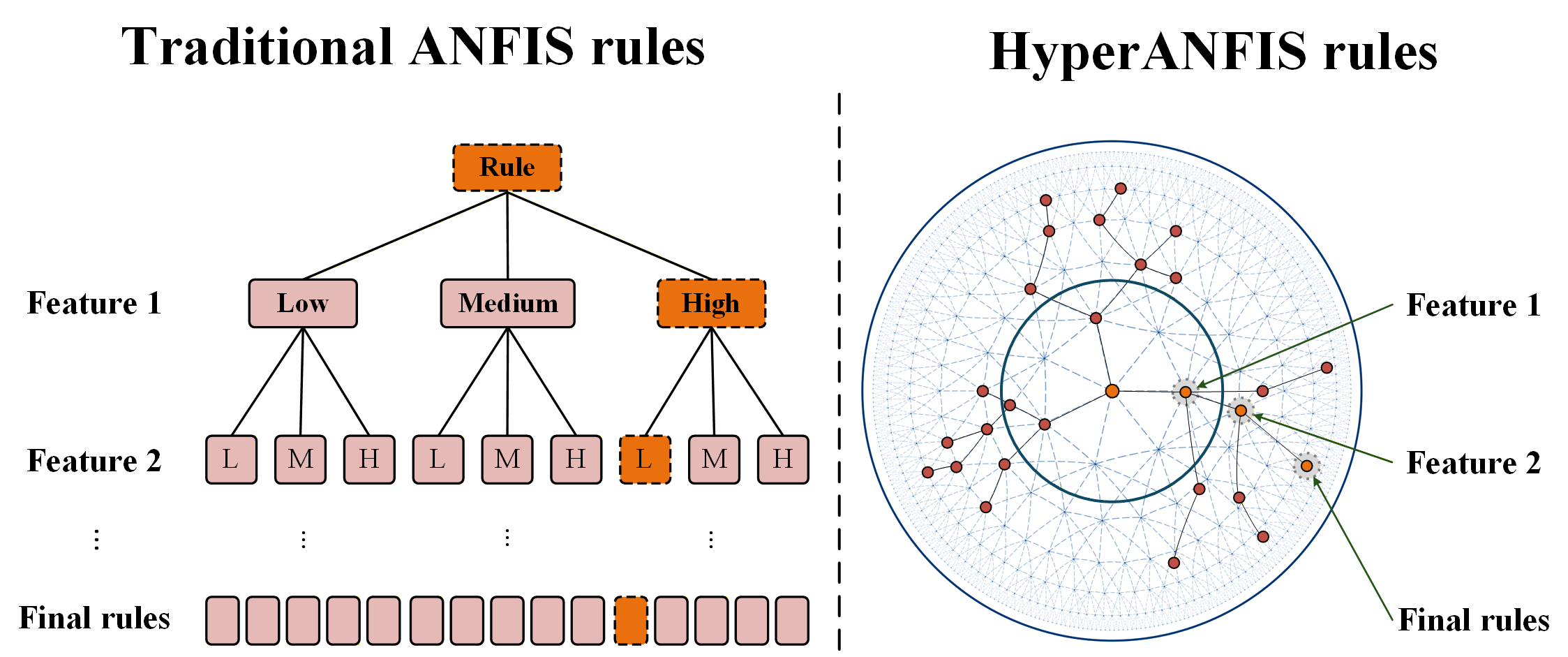}
    \caption{\textbf{Comparison of rule constraints in Euclidean and hyperbolic spaces.} The capacity of Euclidean space is limited, and the rules become severely compressed as the number of features increases. In hyperbolic space, spatial capacity grows continuously, the rules are not constrained by the available space.}
    \label{fig:placeholder}
\end{figure}

This limitation is particularly pronounced when the data or decision regions exhibit hierarchical, tree-like, or highly non-uniform structures. In such cases, Euclidean geometry may provide a weak inductive bias because it does not naturally capture the expanding representational capacity required by increasingly fine-grained branches in a hierarchy \cite{sinha2024learning}. Such structures are widespread across a variety of real-world applications, including healthcare and industrial systems \cite{koo-2024-next}. Therefore, it is important to develop ANFIS models in alternative geometric spaces that can enrich rule representations without sacrificing the transparency of fuzzy IF-THEN reasoning.

Hyperbolic representation learning provides a theoretically motivated geometric alternative for modeling hierarchical, tree-like, and highly non-uniform structure \cite{ganea2018hyperbolic}. Owing to its constant negative curvature, the volume of a metric ball in hyperbolic space grows exponentially with its radius. This property allows a low-dimensional embedding to allocate increasing representational capacity to progressively branching hierarchies \cite{NIPS2017_59dfa2df,ganea2018hyperbolic}. Prior work further suggests that hyperbolic geometry can provide a useful inductive bias when data contain explicit or latent hierarchical and non-uniform structure \cite{sinha2024learning}. 

To this end, we propose Hyperbolic ANFIS (HyperANFIS), a geometry-aware neuro-fuzzy model. HyperANFIS reformulates the ANFIS architecture by exploiting the properties of hyperbolic manifolds. It represents fuzzy rules using global geodesic prototypes, derives rule activation strengths from geodesic distances, and aggregates rule consequents on a hyperbolic manifold. This design preserves the scalar fuzzy semantics and functional roles of ANFIS while providing each rule with an explicit geometric interpretation. Under matched experimental settings, HyperANFIS achieves better predictive performance than its Euclidean ANFIS baseline and produces more credible rule representations.

Our contributions are as follows:
\begin{itemize}
    \item We propose HyperANFIS, a model that learns global IF-THEN rules on a hyperbolic manifold and makes accurate predictions based on these rules.
    \item HyperANFIS unifies rule matching, local consequent construction, and consequent aggregation on a hyperbolic manifold, offering a new perspective on mitigating the limitations of Euclidean geometry in ANFIS.
    \item Experiments on multiple real-world datasets show that HyperANFIS outperforms the compared ANFIS models and produces more credible IF-THEN rules.
\end{itemize}

\begin{figure*}[t!]
    \centering
    \includegraphics[
        width=\textwidth,
        trim=20 110 50 35,
        clip
    ]{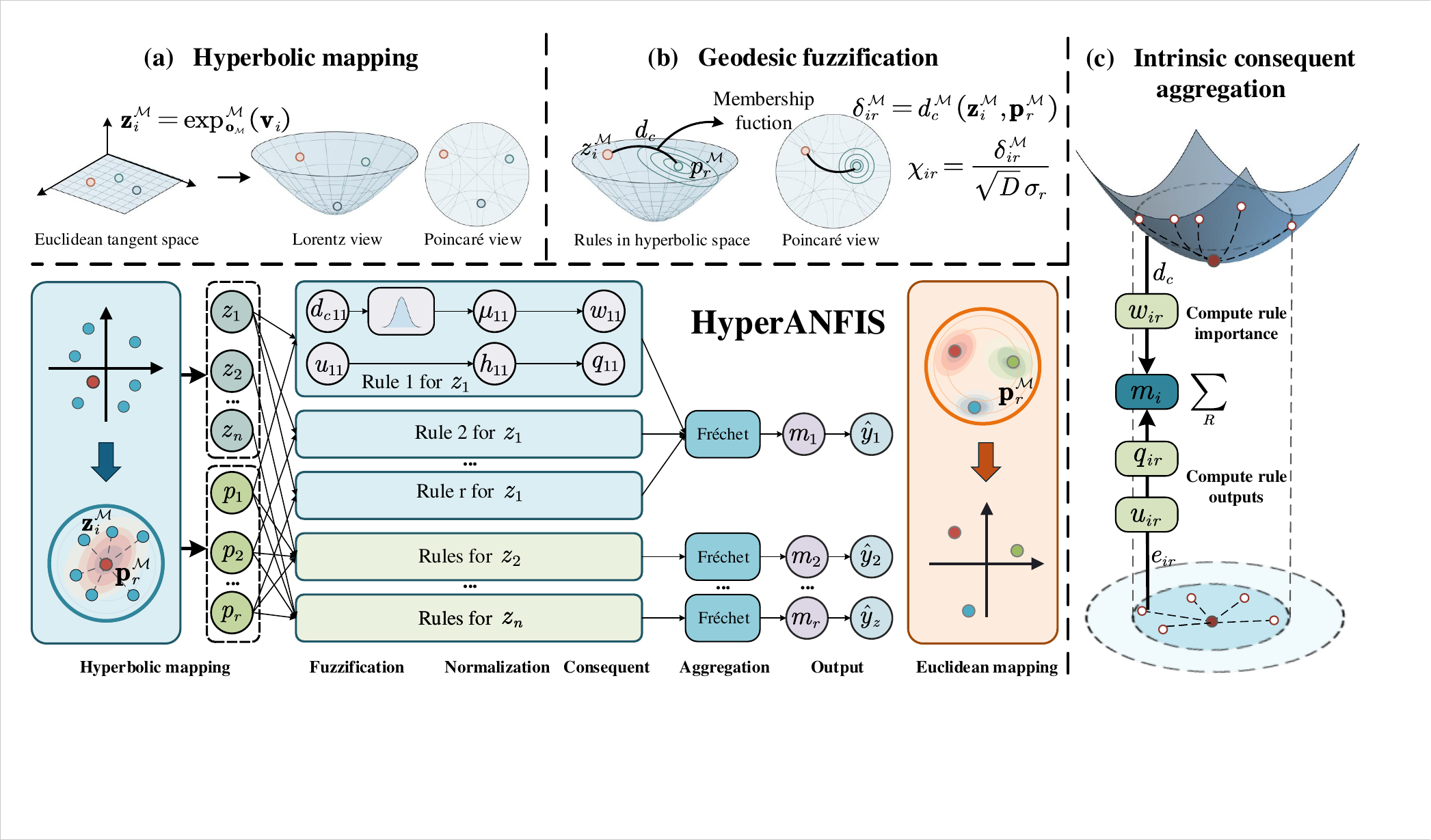}

    \caption{\textbf{Architecture of the HyperANFIS model.} The upper-left panel presents the complete model architecture. (a), (b), and (c) illustrate, respectively, the mapping of samples and rule prototypes into hyperbolic space, the generation of fuzzy rules in hyperbolic space, and the rule-based inference process.}
    \label{fig:hyperanfis_architecture}
\end{figure*}

\section{Related Work}

HyperANFIS is a model that reconstructs ANFIS in hyperbolic space to derive highly interpretable rules; accordingly, research on interpretable machine learning, neuro-fuzzy systems, and hyperbolic models provides the theoretical foundation for this work.

\subsection{Interpretable Machine Learning}

ANFIS combine adaptive parameter learning with Takagi-Sugeno fuzzy IF-THEN inference and have long been widely used in nonlinear approximation, prediction, control, and classification tasks \cite{256541}. Recent studies have explored different strategies to improve the scalability and interpretability of ANFIS. For example, Jin et al. simplify high-dimensional ANFIS by selecting important rules and removing similar ones \cite{jin2024simplification}, while UNFIS learns unstructured fuzzy rules that relax the requirement of evaluating every input variable within each rule \cite{salimibadr2024unfis}. Beyond rule simplification and structural relaxation, recent structure-learning approaches, such as SL-ANFIS-LSTM \cite{su2025sl}, introduce adaptive fuzzy structure optimization to improve the flexibility of ANFIS-based models. More recently, KANFIS further explores a neuro-symbolic formulation based on additive function decomposition and sparse masking to reduce rule complexity and improve interpretability of fuzzy systems \cite{yong2026kanfis}. These studies address important challenges such as rule explosion, redundancy, and antecedent complexity from different perspectives. However, existing approaches mainly focus on rule selection, structural formulation, or functional decomposition, while the geometric organization of rule representations remains largely unexplored. HyperANFIS explores the possibility of realizing ANFIS in hyperbolic space, providing a geometric perspective for improving rule representation.

\subsection{ANFIS and Neuro-Fuzzy Learning}

ANFIS combine adaptive parameter learning with Takagi-Sugeno fuzzy IF-THEN inference and have long been widely used in nonlinear approximation, prediction, control, and classification tasks \cite{256541}. Recent studies on rule-base simplification, redundancy reduction, and the trade-off between predictive performance and interpretability indicate that ANFIS remains an active research area. For example, Jin et al. simplify high-dimensional ANFIS by selecting important rules and removing similar ones \cite{jin2024simplification}, whereas UNFIS learns unstructured fuzzy rules that do not require every input variable to be evaluated within each rule \cite{salimibadr2024unfis}. More recently, KANFIS introduces a neuro-symbolic formulation based on additive function decomposition and sparse masking to reduce rule complexity and improve the interpretability of fuzzy systems \cite{yong2026kanfis}. These studies address important issues such as rule explosion, rule redundancy, and excessively complex antecedents from different perspectives. However, existing approaches mainly focus on rule selection, structural formulation, or functional decomposition, while the geometric organization of rule representations remains largely unexplored. HyperANFIS explores the possibility of realizing ANFIS in hyperbolic space, providing a geometric perspective for improving rule representation.

\subsection{Theoretical Foundations and Empirical Applications of Hyperbolic Geometry}

Hyperbolic geometry is theoretically well suited for representing tree-like and heterogeneous structures. In a space of constant negative curvature, the volume of a geodesic ball grows exponentially with its radius, more closely matching the branching expansion of trees than the polynomial volume growth of Euclidean space. The relationship among negative curvature, hierarchical organization, and heterogeneous networks has been extensively studied in complex-network theory \cite{krioukov2010hyperbolic}. 

In representation learning, Poincaré embeddings have demonstrated that latent hierarchies can be compactly encoded in low-dimensional hyperbolic space \cite{NIPS2017_59dfa2df}; related work on hyperbolic neural networks has provided a systematic formulation of neural computation based on exponential maps, logarithmic maps, and gyrovector operations \cite{ganea2018hyperbolic}, while hyperbolic graph convolutional networks have shown that such operations can preserve the hierarchical and scale-free properties of graph structures during message passing \cite{chami2019hyperbolic}. Taken together, these theoretical results suggest that hyperbolic geometry may improve representation learning when the underlying task involves structured relationships. In addition, some studies have used hyperbolic objectives to encode label hierarchies, thereby reducing distortion in visual representations \cite{sinha2024learning}; others have combined spherical and hyperbolic components to model cyclical homophily and hierarchical social influence in social networks, respectively \cite{iyer2024non}. In visual learning, hyperbolic visual hierarchy mapping supports classification and dense prediction tasks \cite{kwon2024improving}, and hyperbolic distances have also been used in open-world object detection to capture superclass relationships \cite{doan2024hyp}. This line of research has further expanded to multiple modalities, with hyperbolic representations applied to compositional vision-language learning and genomic sequence modeling \cite{pal2025compositional,khan2025hyperbolic}. Recent work has also proposed low-distortion, GPU-compatible hyperbolic tree embeddings \cite{van2025low} and investigated hyperbolic embeddings of supervised models, including decision trees \cite{nock2024hyperbolic}; continuous hyperbolic latent representations have been introduced for hierarchical open-vocabulary 3D scenes \cite{weijler2026openhype}, and hyperbolic representation learning has been explored for reasoning over the hierarchical structure of program syntax \cite{zhou2026iclr-natural}. 

However, hyperbolization methods have not been widely studied in intrinsically interpretable models. To the best of our knowledge, this work is the first to reconstruct ANFIS in hyperbolic space, and our experiments demonstrate its effectiveness.

Overall, these studies suggest that reconstructing ANFIS in hyperbolic space has a solid research basis, promising application potential, and methodological feasibility. Building on this foundation, this work further investigates the construction and performance of HyperANFIS, with the aim of providing new insights into the structural refinement of ANFIS and the application of hyperbolic space in intrinsically interpretable models.

\section{Methodology}
\label{sec:methodology}

As shown in Figure~\ref{fig:hyperanfis_architecture}, HyperANFIS extends the neuro-fuzzy inference process to hyperbolic space, leveraging negatively curved geometry to represent fuzzy rules and perform inference. This design improves the model's representational capacity and supports more discriminative, stable, and auditable fuzzy rules. This section details the architecture of HyperANFIS and its key methodological components.

\subsection{Overall Architecture of HyperANFIS}
\label{sec:overall_architecture}

As shown in Figure~\ref{fig:hyperanfis_architecture}, HyperANFIS is organized as five functional layers. Given a standardized input, the first layer maps the input and all rule centers from a shared origin tangent space to the selected hyperbolic representation, and then evaluates their intrinsic geodesic distances. The second layer converts each sample-to-rule distance into one scalar radial membership. The third layer normalizes the log-memberships across rules to obtain firing strengths. The fourth layer constructs a rule-specific consequent from the intrinsic local coordinates of the sample. The fifth layer maps these consequents to the output manifold, aggregates them through a firing-weighted Fr\'echet mean, and decodes the aggregate for classification or regression. All trainable parameters are optimized jointly through the resulting differentiable computation graph.

\subsection{Hyperbolic Mapping and Rule Pullback}
\label{sec:hyperbolic_mapping}

Let $\widehat{\mathbf{x}}_i\in\mathbf{R}^{D}$ denote the standardized input of sample $i$, and let $\mathbf{a}_r\in\mathbf{R}^{D}$ be the trainable tangent center of rule $r$. With an input scale $s>0$ and tangent bound $\tau>0$, HyperANFIS forms
\begin{equation}
\mathbf{v}_i=\mathrm{clip}_{\tau}(s\widehat{\mathbf{x}}_i),
\qquad
\overline{\mathbf{a}}_r=\mathrm{clip}_{\tau}(\mathbf{a}_r),
\label{eq:tangent_parameterization}
\end{equation}
where
\begin{equation}
\mathrm{clip}_{\tau}(\mathbf{v})
=
\mathbf{v}\min\left(1,\frac{\tau}{\|\mathbf{v}\|}\right).
\label{eq:tangent_clipping}
\end{equation}
The clipping is radial, so it preserves direction while bounding the represented geodesic radius. Define $\varrho(\mathbf{v})=\sqrt{c}\|\mathbf{v}\|$, where $-c$ is the sectional curvature and $c>0$.

For the Lorentz representation, the origin is $\mathbf{o}_{\mathrm{L}}=(c^{-1/2},0,\ldots,0)$. The origin exponential map is
\begin{equation}
\exp_{\mathbf{o}_{\mathrm{L}}}^{\mathrm{L}}(\mathbf{v})
=
\left(
c^{-1/2}\cosh\left(\varrho(\mathbf{v})\right),
\frac{\sinh\left(\varrho(\mathbf{v})\right)}
{\varrho(\mathbf{v})}\mathbf{v}
\right).
\label{eq:lorentz_exp}
\end{equation}
For the Poincar\'e representation, the origin is $\mathbf{o}_{\mathrm{P}}=\mathbf{0}$. Under the shared tangent convention used by HyperANFIS, its exponential map is
\begin{equation}
\exp_{\mathbf{o}_{\mathrm{P}}}^{\mathrm{P}}(\mathbf{v})
=
\frac{\tanh\left(\varrho(\mathbf{v})/2\right)}
{\varrho(\mathbf{v})}\mathbf{v}.
\label{eq:poincare_exp}
\end{equation}
These maps follow the standard negative-curvature constructions used in hyperbolic representation learning \cite{ganea2018hyperbolic,nickel2018lorentz}. HyperANFIS selects either $\mathcal{M}=\mathrm{L}$ or $\mathcal{M}=\mathrm{P}$ and constructs
\begin{equation}
\mathbf{z}_i^{\mathcal{M}}
=\exp_{\mathbf{o}_{\mathcal{M}}}^{\mathcal{M}}(\mathbf{v}_i),
\qquad
\mathbf{p}_r^{\mathcal{M}}
=\exp_{\mathbf{o}_{\mathcal{M}}}^{\mathcal{M}}(\overline{\mathbf{a}}_r).
\label{eq:sample_rule_embedding}
\end{equation}
The two representations are alternative coordinate choices and share the same trainable tangent centers.

For rule inspection, let $x_{ij}^{\mathrm{raw}}$ be a raw feature with training-set location $\nu_j$ and scale $\zeta_j>0$. The preprocessing and inverse center map are
\begin{equation}
\begin{array}{rcl}
\widehat{x}_{ij} &=& (x_{ij}^{\mathrm{raw}}-\nu_j)/\zeta_j,\\
\widetilde{a}_{rj} &=& \nu_j+\zeta_j\overline{a}_{rj}/s.
\end{array}
\label{eq:raw_rule_pullback}
\end{equation}
Thus, $\widetilde{\mathbf{a}}_r$ is a readable representative center in the original feature coordinates. It is not a coordinate-wise Euclidean rule boundary. Exact rule evaluation still replays preprocessing, clipping, hyperbolic mapping, and intrinsic distance computation, as detailed in \textbf{Appendix A}.

\subsection{Geodesic Fuzzy Rule Construction}
\label{sec:geodesic_rules}

HyperANFIS defines each antecedent as one geodesic region around a global rule prototype. For Lorentz points $\mathbf{z},\mathbf{p}\in\mathcal{H}_c^D$, define
\begin{equation}
\left\langle\mathbf{z},\mathbf{p}\right\rangle_{\mathrm{L}}
=-z_0p_0+\sum_{j=1}^{D}z_jp_j.
\label{eq:lorentz_inner}
\end{equation}
The corresponding intrinsic distance is
\begin{equation}
d_c^{\mathrm{L}}(\mathbf{z},\mathbf{p})
=c^{-1/2}\mathrm{arcosh}
\left(-c\left\langle\mathbf{z},\mathbf{p}\right\rangle_{\mathrm{L}}\right).
\label{eq:lorentz_distance}
\end{equation}
For Poincar\'e points $\mathbf{z},\mathbf{p}\in\mathcal{B}_c^D$, let $\Delta_c(\mathbf{q})=1-c\|\mathbf{q}\|^2$. Their intrinsic distance is
\begin{equation}
d_c^{\mathrm{P}}(\mathbf{z},\mathbf{p})
=c^{-1/2}\mathrm{arcosh}
\left(
1+\frac{2c\|\mathbf{z}-\mathbf{p}\|^2}
{\Delta_c(\mathbf{z})\Delta_c(\mathbf{p})}
\right).
\label{eq:poincare_distance}
\end{equation}
For either representation, the sample-to-rule distance is
\begin{equation}
\delta_{ir}^{\mathcal{M}}
=d_c^{\mathcal{M}}
\left(\mathbf{z}_i^{\mathcal{M}},\mathbf{p}_r^{\mathcal{M}}\right).
\label{eq:intrinsic_rule_distance}
\end{equation}

Each rule has one learnable intrinsic scale. HyperANFIS parameterizes it with an unconstrained logit $\beta_r$:
\begin{equation}
\sigma_r
=
\sigma_{\min}
+(\sigma_{\max}-\sigma_{\min})\mathrm{sigmoid}(\beta_r).
\label{eq:bounded_rule_scale}
\end{equation}
The dimension-normalized distance and Gaussian log-membership are
\begin{equation}
\chi_{ir}=\frac{\delta_{ir}^{\mathcal{M}}}{\sqrt{D}\sigma_r},
\qquad
\ell_{ir}=-\frac{1}{2}\chi_{ir}^{2},
\qquad
\mu_{ir}=\exp(\ell_{ir}).
\label{eq:gaussian_membership}
\end{equation}
Consequently, $\sqrt{D}\sigma_r$ is the geodesic radius at which $\mu_{ir}=\exp(-1/2)$. The implementation also supports a generalized Bell kernel, whose exact form is given in \textbf{Appendix B}.

The third layer normalizes the log-memberships directly:
\begin{equation}
\overline{w}_{ir}
=
\frac{\exp(\ell_{ir})}
{\sum_{q=1}^{R}\exp(\ell_{iq})}.
\label{eq:normalized_firing}
\end{equation}
This construction yields one scalar activation for each sample-rule pair. It therefore represents a global geodesic fuzzy rule rather than a product of independent coordinate-wise memberships.

\subsection{Intrinsic Consequent Aggregation}
\label{sec:intrinsic_consequents}

Let $H=\max(2,O)$ be the output-manifold dimension, where $O$ is the task output dimension. HyperANFIS first expresses sample $i$ relative to rule prototype $r$ through the intrinsic local coordinate
\begin{equation}
\mathbf{e}_{ir}
=
\left[
\mathcal{T}_{\mathbf{p}_r^{\mathcal{M}}\rightarrow\mathbf{o}_{\mathcal{M}}}
\left(
\log_{\mathbf{p}_r^{\mathcal{M}}}^{\mathcal{M}}
(\mathbf{z}_i^{\mathcal{M}})
\right)
\right]_{\mathrm{sp}},
\label{eq:local_rule_coordinate}
\end{equation}
where $\mathcal{T}$ denotes parallel transport and $[\cdot]_{\mathrm{sp}}$ retains the $D$ spatial tangent coordinates. A first-order rule consequent is
\begin{equation}
\begin{array}{rcl}
\mathbf{u}_{ir}
&=&
\mathrm{clip}_{\tau}
\left(\mathbf{b}_r+\mathbf{W}_r\mathbf{e}_{ir}\right),\\
\mathbf{q}_{ir}^{\mathcal{M}}
&=&
\exp_{\mathbf{o}_{\mathcal{M}}}^{\mathcal{M}}(\mathbf{u}_{ir}).
\end{array}
\label{eq:rule_consequent}
\end{equation}
where $\mathbf{b}_r\in\mathbf{R}^{H}$ and $\mathbf{W}_r\in\mathbf{R}^{H\times D}$. The zero-order variant omits $\mathbf{W}_r\mathbf{e}_{ir}$.

\begin{table*}[t!]
\centering

\resizebox{\textwidth}{!}{
\begin{tabular}{@{}l*{15}{c}@{}}

\toprule
\textbf{Model}
& \multicolumn{3}{c}{\textbf{Spambase}}
& \multicolumn{3}{c}{\textbf{Car}}
& \multicolumn{3}{c}{\textbf{Zoo}}
& \multicolumn{3}{c}{\textbf{WDBC}}
& \multicolumn{3}{c}{\textbf{NSL-KDD}}
\\

\cmidrule(lr){2-4}
\cmidrule(lr){5-7}
\cmidrule(lr){8-10}
\cmidrule(lr){11-13}
\cmidrule(lr){14-16}

& Acc. & F1 & Recall
& Acc. & F1 & Recall
& Acc. & F1 & Recall
& Acc. & F1 & Recall
& Acc. & F1 & Recall
\\

\midrule

FSRE-AdaTSK
& 0.7742 & 0.7544 & 0.7477
& 0.5896 & 0.2619 & 0.2811
& 0.3810 & 0.2379 & 0.3095
& 0.7544 & 0.7304 & 0.7262
& 0.7192 & 0.4235 & 0.4384
\\

FCM-ANFIS
& 0.9229 & 0.9198 & 0.9229
& 0.8815 & 0.8062 & 0.9076
& 0.7142 & 0.6920 & 0.7797
& 0.9035 & 0.8892 & 0.9712
& 0.7932 & 0.5878 & 0.5910
\\

IT2-ANFIS
& 0.9121 & 0.9088 & 0.9130
& 0.9104 & 0.8469 & 0.9086
& 0.8095 & 0.5946 & 0.6607
& 0.9123 & 0.9037 & 0.8958
& 0.7622 & 0.5856 & 0.5842
\\

PSO-ANFIS
& 0.9044 & 0.8996 & 0.8985
& 0.8150 & 0.6517 & 0.9364
& 0.9047 & 0.7809 & 0.8095
& 0.8947 & 0.8889 & 0.8968
& 0.7620 & 0.5602 & 0.5485
\\

ANFIS
& 0.9120 & 0.9080 & 0.9075
& 0.8998 & 0.8277 & 0.8731
& 0.8413 & 0.6875 & 0.7500
& 0.9152 & 0.9067 & 0.8958
& 0.7798 & 0.5952 & 0.5935
\\

\textbf{HyperANFIS (Ours)}
& \textbf{0.9251} & \textbf{0.9215} & \textbf{0.9251}
& \textbf{0.9181} & \textbf{0.8827} & \textbf{0.9449}
& \textbf{0.9524} & \textbf{0.8186} & \textbf{0.8571}
& \textbf{0.9854} & \textbf{0.9843} & \textbf{0.9835}
& \textbf{0.8038} & \textbf{0.6709} & \textbf{0.6281}
\\

\bottomrule
\end{tabular}
}

\caption{\textbf{Performance comparison of HyperANFIS with conventional ANFIS
and other ANFIS variants.} The best available result for each metric on each
dataset is shown in bold. HyperANFIS achieved the best performance in all experimental results.}

\label{tab:main_classification_results}
\end{table*}

The fifth layer aggregates the manifold-valued consequents through their weighted Fr\'echet mean \cite{lou2020frechet}:
\begin{equation}
\mathbf{m}_i^{\mathcal{M}}
=
\mathop{\mathrm{argmin}}_{\mathbf{m}\in\mathcal{M}}
\sum_{r=1}^{R}\overline{w}_{ir}
d_c^{\mathcal{M}}(\mathbf{m},\mathbf{q}_{ir}^{\mathcal{M}})^2,
\label{eq:frechet_aggregation}
\end{equation}
HyperANFIS computes this mean by a fixed number of differentiable Karcher refinements. For classification, trainable class tangents $\mathbf{g}_k$ define class prototypes $\mathbf{c}_k^{\mathcal{M}}=\exp_{\mathbf{o}_{\mathcal{M}}}^{\mathcal{M}}(\mathrm{clip}_{\tau}(\mathbf{g}_k))$, and the class logits are
\begin{equation}
s_{ik}
=
-d_c^{\mathcal{M}}
(\mathbf{m}_i^{\mathcal{M}},\mathbf{c}_k^{\mathcal{M}})^2.
\label{eq:classification_logits}
\end{equation}
For regression, the prediction is obtained in the origin tangent chart:
\begin{equation}
\widehat{\mathbf{y}}_i
=
\left[
\log_{\mathbf{o}_{\mathcal{M}}}^{\mathcal{M}}
(\mathbf{m}_i^{\mathcal{M}})
\right]_{1:O}.
\label{eq:regression_decoder}
\end{equation}

\subsection{Forward Computation}
\label{sec:overall_forward}

For a standardized input $\mathbf{x}_i$, the complete HyperANFIS forward computation is summarized as

\begin{equation}
\begin{array}{rcl}
\left\{
(\overline{w}_{ir},\mathbf{q}_{ir}^{\mathcal{M}})
\right\}_{r=1}^{R}
&=&
\mathcal{R}_{\Theta,\mathcal{M}}(\mathbf{x}_i),\\[2pt]
\mathbf{m}_i^{\mathcal{M}}
&=&
\mathop{\mathrm{argmin}}\limits_{\mathbf{m}\in\mathcal{M}}
\displaystyle\sum_{r=1}^{R}
\overline{w}_{ir}
d_c^{\mathcal{M}}
(\mathbf{m},\mathbf{q}_{ir}^{\mathcal{M}})^2,\\[2pt]
\widehat{\mathbf{y}}_i
&=&
\mathcal{D}(\mathbf{m}_i^{\mathcal{M}}).
\end{array}
\label{eq:overall_forward}
\end{equation}
Here, $\mathcal{R}_{\Theta,\mathcal{M}}$ denotes hyperbolic rule
inference and produces the normalized firing strength
$\overline{w}_{ir}$ and intrinsic consequent
$\mathbf{q}_{ir}^{\mathcal{M}}$ for each rule. These consequents are
then aggregated through their activation-weighted Fr\'echet mean, and
$\mathcal{D}$ decodes the resulting manifold point into the task
prediction. Detailed definitions and optimization procedures are
provided in the \textbf{Appendix C}.

\section{Experiments}

To systematically evaluate whether hyperbolic geometric modeling improves the predictive performance and rule interpretability of ANFIS, we compare HyperANFIS with conventional ANFIS and other representative ANFIS variants across multiple datasets. The selected datasets cover several application domains, including healthcare and finance, and vary in complexity, thereby providing a broad experimental basis for model evaluation. In addition, we use the Wisconsin Diagnostic Breast Cancer (WDBC) dataset as a case study and analyze the generated rules from multiple perspectives, providing further evidence of the improved rule interpretability achieved by HyperANFIS.

\subsection{Comparison Experiments}

We compared HyperANFIS with conventional ANFIS, which served as the primary Euclidean baseline, and four representative neuro-fuzzy models. Conventional ANFIS adopts a dimension-wise Takagi-Sugeno inference structure and learns its antecedent and consequent parameters from the training data \cite{zhang2024scikitanfis}. FSRE-AdaTSK integrates feature selection, rule extraction, and parameter fine-tuning within an adaptive TSK fuzzy system \cite{xue2023fsre}. FCM-ANFIS uses fuzzy c-means clustering to initialize fuzzy partitions and rule prototypes \cite{knaiber2023bayesian}. In our implementation, these prototypes were subsequently refined through gradient-based optimization. IT2-ANFIS represents uncertainty in rule membership by introducing lower and upper interval type-2 membership bounds into its coordinate-wise rule antecedents \cite{zand2024hybrid}. In our implementation, a midpoint approximation was obtained by averaging the lower and upper bounds of the corresponding rule firing strengths. Finally, PSO-ANFIS uses particle swarm optimization to optimize the trainable parameters of ANFIS \cite{vesovic2024hybrid}. In our implementation, PSO provided a supervised warm start for all trainable parameters, followed by a common Adam-based refinement stage.

The five datasets cover email spam detection (Spambase), vehicle
acceptability evaluation (Car), animal category classification (Zoo),
breast cancer diagnosis (WDBC), and network intrusion detection
(NSL-KDD). They encompass both binary and multiclass classification
tasks and differ in sample size, feature dimensionality, feature
characteristics, and class structure, providing a diverse test bed
for evaluating model performance across application domains and data
distributions. For each random seed, all methods used consistent data
preprocessing, data partitions, and evaluation metrics, and the
reported results were averaged over five predefined random seeds.

As shown in Table~\ref{tab:main_classification_results}, HyperANFIS outperforms conventional ANFIS and the representative neuro-fuzzy baselines on all five datasets. Compared with conventional ANFIS, HyperANFIS improves the average accuracy, Macro-F1 score, and Recall by 0.0473, 0.0706, and 0.0638, respectively. As discussed in the following sections, HyperANFIS also yields more reliable predictions than the comparison models in terms of the output rule attributes. These results demonstrate that geometry-aware rule inference can substantially enhance the predictive capability of ANFIS and indicate that hyperbolic computation is a promising direction for further improving ANFIS and its variants.

These improvements are particularly pronounced on Zoo and WDBC. On the Zoo dataset, HyperANFIS improves accuracy, Macro-F1, and Recall by 11.11, 13.11, and 0.1071 percentage points, respectively; on WDBC, the corresponding improvements reach 7.02, 7.76, and 0.0877 percentage points. Zoo exhibits a natural hierarchical taxonomic structure, whereas WDBC contains complex and nonuniform diagnostic relationships among its features. These consistent improvements on the two datasets provide empirical support for the effectiveness of negative-curvature geometry in organizing nonuniform and hierarchical rule relationships, thereby leading to performance gains.

\begin{table*}[t]
\centering
\small
\setlength{\tabcolsep}{3pt}
\renewcommand{\arraystretch}{1.08}

\begin{tabular}{@{}p{0.07\textwidth}p{0.40\textwidth}p{0.50\textwidth}@{}}
\hline
\textbf{Rule} &
\textbf{Projected, interpretable IF-THEN rule} &
\textbf{Morphological evidence} \\
\hline

\textbf{R4} &
\textbf{IF} \textit{mean compactness} \textbf{is HIGH}
($\widetilde c=+3.69$), \textit{mean concavity} \textbf{is HIGH}
($\widetilde c=+3.65$), \textit{mean concave points} \textbf{is HIGH}
($\widetilde c=+3.77$), and \textit{worst concave points} \textbf{is HIGH}
($\widetilde c=+3.84$).
\newline\textbf{THEN} the rule-level conclusion is
\textbf{malignant (M)}. \textit{Dominant coverage: 26.3\%.}
&
High compactness, concavity, and concave-point counts describe pronounced
nuclear-boundary indentation and irregularity. Direct breast-aspirate
morphometry found compactness and concave-point counts to be significantly
higher in ductal carcinoma than in benign lesions
\cite{narasimha2013significance}.
\\[3pt]

\textbf{R10} &
\textbf{IF} \textit{mean radius} \textbf{is LOW}
($\widetilde c=-2.24$), \textit{mean concavity} \textbf{is LOW}
($\widetilde c=-1.98$), \textit{mean concave points} \textbf{is LOW}
($\widetilde c=-2.18$), and \textit{worst radius} \textbf{is LOW}
($\widetilde c=-2.37$).
\newline\textbf{THEN} the rule-level conclusion is
\textbf{benign (B)}. \textit{Dominant coverage: 21.1\%.}
&
Small nuclei with low concavity and few concave points form a compact,
comparatively smooth nuclear phenotype. Direct breast-aspirate morphometry
reported lower nuclear size, compactness, and concave-point counts in benign
lesions than in ductal carcinoma
\cite{narasimha2013significance}.
\\[3pt]

\textbf{R2} &
\textbf{IF} \textit{mean radius} \textbf{is LOW}
($\widetilde c=-1.11$), \textit{mean area} \textbf{is LOW}
($\widetilde c=-1.21$), \textit{worst radius} \textbf{is LOW}
($\widetilde c=-1.32$), and \textit{worst area} \textbf{is LOW}
($\widetilde c=-0.70$).
\newline\textbf{THEN} the rule-level conclusion is
\textbf{benign (B)}. \textit{Dominant coverage: 14.9\%.}
&
Low mean and worst-case radius and area indicate that both typical and extreme
nuclear sizes remain small. This size-dominant pattern agrees with cytological
evidence showing lower nuclear area, diameter, and perimeter in benign breast
aspirates
\cite{pandian2021image}.
\\[3pt]

\textbf{R6} &
\textbf{IF} \textit{mean concavity} \textbf{is LOW}
($\widetilde c=-2.06$), \textit{radius SE} \textbf{is LOW}
($\widetilde c=-2.23$), \textit{perimeter SE} \textbf{is LOW}
($\widetilde c=-2.12$), and \textit{worst radius} \textbf{is LOW}
($\widetilde c=-2.37$).
\newline\textbf{THEN} the rule-level conclusion is
\textbf{benign (B)}. \textit{Dominant coverage: 12.3\%.}
&
Low concavity and worst radius support a compact, regular nuclear phenotype;
low radius and perimeter SE indicate limited within-sample size dispersion.
This direction agrees with the greater nuclear size and size dispersion
reported in malignant aspirates
\cite{pandian2021image}.
\\[3pt]

\textbf{R9} &
\textbf{IF} \textit{mean radius} \textbf{is HIGH}
($\widetilde c=+1.60$), \textit{mean area} \textbf{is HIGH}
($\widetilde c=+1.24$), \textit{worst perimeter} \textbf{is HIGH}
($\widetilde c=+2.05$), and \textit{worst area} \textbf{is HIGH}
($\widetilde c=+1.15$).
\newline\textbf{THEN} the rule-level conclusion is
\textbf{malignant (M)}. \textit{Dominant coverage: 12.3\%.}
&
High mean size and high extreme perimeter and area jointly capture nuclear
enlargement and exceptionally large nuclei. Breast-aspirate morphometry
similarly reported substantially greater nuclear area, perimeter, and diameter
in malignant lesions
\cite{pandian2021image}.
\\[3pt]

\textbf{R12} &
\textbf{IF} \textit{mean radius} \textbf{is LOW}
($\widetilde c=-2.00$), \textit{mean area} \textbf{is LOW}
($\widetilde c=-1.15$), \textit{radius SE} \textbf{is HIGH}
($\widetilde c=+1.29$), and \textit{worst radius} \textbf{is LOW}
($\widetilde c=-1.66$).
\newline\textbf{THEN} the rule-level conclusion is
\textbf{benign (B)}. \textit{Dominant coverage: 6.1\%.}
&
This mixed phenotype combines small typical and extreme nuclear sizes with
greater radius variation. Small nuclear size supports the benign consequent,
while single-cell morphometry confirms that measurable variation in nuclear
size and contour can also occur within benign breast disease
\cite{abubakar2025spatially}.
\\

\hline
\end{tabular}

\caption{\textbf{Selected HyperANFIS rules learned on the WDBC dataset.}
Each row represents a single IF-THEN rule together with the corresponding
support from the literature. An individual rule is not the prediction for a
single sample; rather, its firing strength is influenced by its dominant
coverage.}
\label{tab:readable_hyperanfis_rules}

\end{table*}

\subsection{Interpretability Analysis}

To evaluate whether hyperbolic geometry improves the rule interpretability of ANFIS, we compared HyperANFIS with classical ANFIS on the Wisconsin Diagnostic Breast Cancer (WDBC) dataset and analyzed representative IF-THEN rules. WDBC contains 569 instances and 30 real-valued features extracted from digitized images of fine-needle aspirates of breast masses. These features describe nuclear size, texture, and contour morphology, providing explicit cytomorphological semantics for evaluating rule utilization, distinctiveness, and morphological consistency.

As shown in Table~\ref{tab:readable_hyperanfis_rules}, HyperANFIS preserves the ability of classical ANFIS to express its inference process through interpretable IF-THEN rules. For readability, the table reports the six rules with the highest dominant coverage on the validation set and retains only three to four representative features for each rule, while the omitted features remain part of the complete antecedent. The learned rules associate malignancy with increased nuclear size and contour irregularity, whereas benign patterns are generally characterized by smaller nuclear size and lower concavity. These patterns are consistent with established findings in breast nuclear morphometry.

\begin{figure*}[t!]
    \centering
    \includegraphics[
    width=\textwidth,
        trim=10 30 5 10,
        clip
    ]{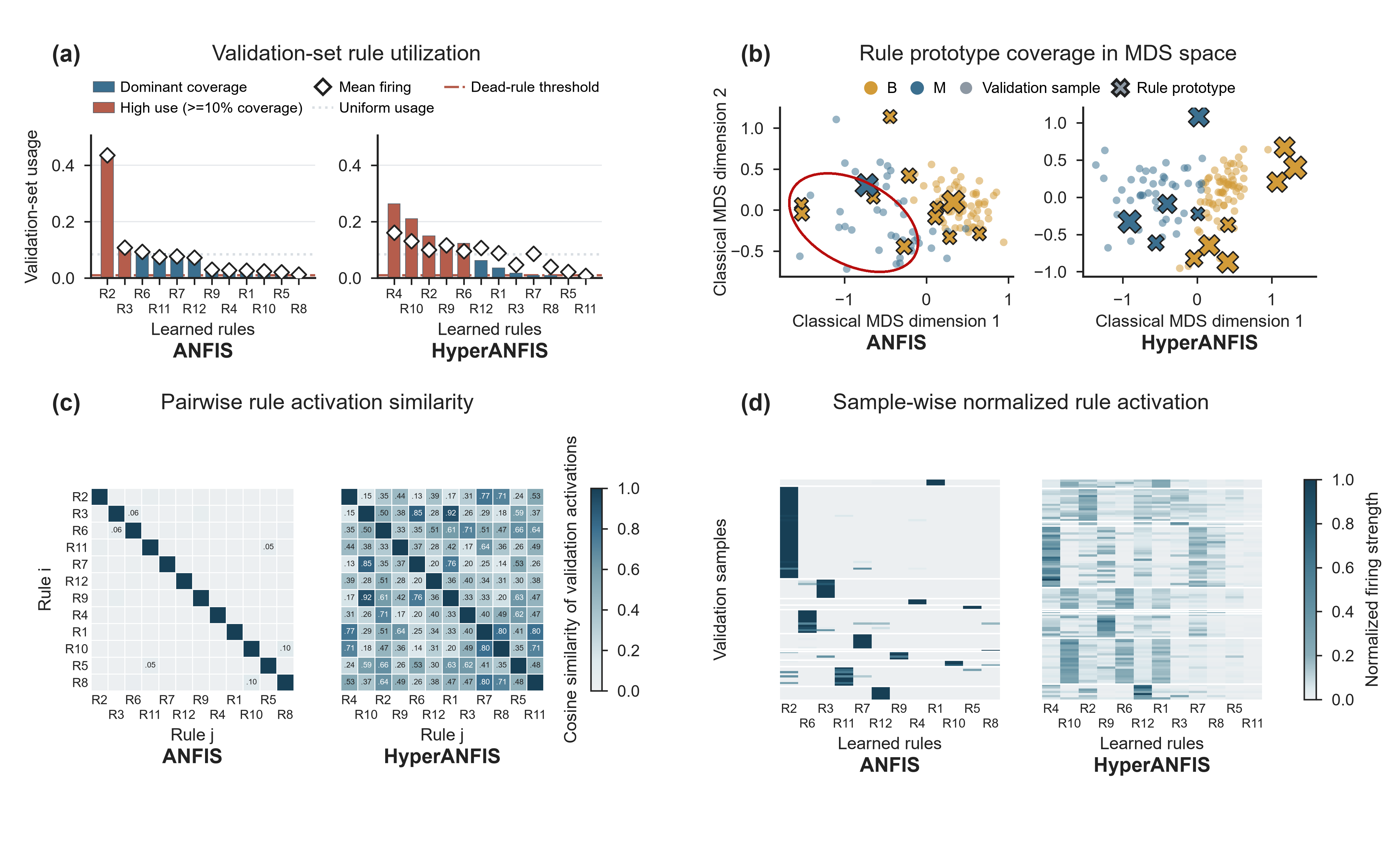}

    \caption{\textbf{Comparison of the interpretability of ANFIS and
    HyperANFIS on the WDBC dataset.} The ANFIS results are shown on the left,
    whereas the HyperANFIS results are shown on the right. (a) shows the
    frequency with which each rule is used for prediction, (b) visualizes the relationship between the rules and the dataset, with red circles marking the centers of rules associated with severe errors, (c) shows the correlations among the rules, and (d) illustrates the contribution of each rule to each individual prediction.}
    \label{fig:bcw_rule_analysis}
\end{figure*}

\subsection{Comparison with Classical ANFIS}

We compare the rule characteristics learned by conventional ANFIS and HyperANFIS on the WDBC dataset, which has a complex feature space and strong intrinsic logical structure. As shown in Figure~\ref{fig:bcw_rule_analysis}, the interpretability of conventional ANFIS deteriorates substantially on this dataset, where the feature complexity and underlying relational structure make rule learning challenging. In contrast, HyperANFIS preserves a high level of interpretability, revealing a clear distinction between the two models. This experiment was conducted with a fixed seed under fair comparison conditions.

In Figure 3(a), classical ANFIS assigns 42.98\% of the validation samples to R2 as their dominant rule, whereas the dominant coverage of the second most frequently selected rule drops to 10.53\%. In contrast, HyperANFIS distributes dominant coverage across five principal rules: R4 (26.32\%), R10 (21.05\%), R2 (14.91\%), R6 (12.28\%), and R9 (12.28\%). The remaining rules also participate in inference through nonzero mean firing strengths. Thus, HyperANFIS transforms the single-rule-dominated utilization pattern of classical ANFIS into a broader and more clearly organized multi-rule participation pattern.
Figure 3(b) further demonstrates this improvement at the rule-prototype level. Classical ANFIS learns 11 benign prototypes but only one malignant prototype, resulting in a substantial imbalance in class representation. In contrast, HyperANFIS learns seven benign and five malignant prototypes, whose distribution is more coherent. As indicated by the red circles in the figure, some rule prototypes are also clearly mislearned. 
Figure 3(c) shows that the mean off-diagonal cosine similarity between rule activation vectors increases from 0.006 to 0.432. The rules learned by classical ANFIS are activated in an almost completely mutually exclusive manner, which appears difficult to reconcile with the principles of fuzzy inference. HyperANFIS, in contrast, establishes correlated yet differentiated relationships among its rules, thereby better conforming to the gradual evidence-combination process underlying fuzzy reasoning. 
Figure 3(d) provides direct sample-level evidence: the mean entropy-derived effective number of active rules increases from 1.14 for classical ANFIS to 6.67 for HyperANFIS. Classical ANFIS consequently exhibits an almost winner-take-all inference pattern dominated by a single rule, whereas HyperANFIS uses one or more principal rules to guide the inference, with the remaining rules providing supplementary evidence.

These results indicate that hyperbolic geometry enables the learned IF-THEN rules to provide more complete class representations, stronger inter-rule cooperation, and a more trustworthy inference process. Together with the accuracy results reported in Table~\ref{tab:main_classification_results}, the findings show that HyperANFIS achieves higher accuracy and more trustworthy predictions while producing higher-quality rules.

\section{Conclusion}

This work investigated how the geometric properties of hyperbolic space can improve ANFIS, an intrinsically interpretable model. By reformulating rule prototypes, rule activation, and consequent aggregation in hyperbolic space while preserving the original IF-THEN semantics, HyperANFIS consistently outperformed classical ANFIS and representative ANFIS variants in accuracy and Macro-F1 across the five evaluated datasets. The WDBC analysis further demonstrated that HyperANFIS learned a more class-diverse, cooperative, and auditable rule system than classical ANFIS. The significance of this work therefore extends beyond the geometric reformulation and performance improvement of classical ANFIS. It provides a new geometric perspective for improving other ANFIS-family models and, more broadly, intrinsically interpretable models.

\bigskip

\bibliography{aaai2027}


\clearpage

\appendix

\section*{Appendix}

This supplement provides additional geometric definitions, model variants,
optimization details, theoretical analysis, and implementation settings for
HyperANFIS.

\section{A Hyperbolic Geometry}
\label{app:geometry_details}

We define the Lorentz and Poincar\'e operations under a shared tangent-space
parameterization. Throughout, $c>0$ denotes the curvature magnitude, and the
sectional curvature is $-c$. Input scaling, projection, and finite-precision
safeguards are specified in
Appendix~\ref{app:implementation_details}.

\subsection{A.1 Tangent-Space Parameterization}
\label{app:tangent_parameterization}

Let $\widehat{\mathbf{x}}_i\in\mathbb{R}^{D}$ denote a standardized input and
let $\mathbf{a}_r\in\mathbb{R}^{D}$ denote the trainable tangent-space center
of rule $r$. HyperANFIS parameterizes samples and rule centers as
\begin{equation}
\mathbf{v}_i
=
\mathrm{clip}_{\tau}(s\widehat{\mathbf{x}}_i),
\qquad
\overline{\mathbf{a}}_r
=
\mathrm{clip}_{\tau}(\mathbf{a}_r),
\label{eq:app_tangent_parameterization}
\end{equation}
where $s>0$ is the input scale and $\tau>0$ is the maximum tangent norm. The
clipping operator is
\begin{equation}
\mathrm{clip}_{\tau}(\mathbf{v})
=
\begin{cases}
\mathbf{0}, & \|\mathbf{v}\|=0,\\[2pt]
\mathbf{v}\min\left(1,\dfrac{\tau}{\|\mathbf{v}\|}\right),
& \|\mathbf{v}\|>0.
\end{cases}
\label{eq:app_tangent_clipping}
\end{equation}
Radial clipping preserves the direction of each nonzero tangent vector and
bounds the corresponding geodesic radius. The definition at zero gives the
continuous extension. The data-dependent scale $s$ is specified in
Appendix~\ref{app:input_scaling_and_initialization}.

\subsection{A.2 Lorentz Model}
\label{app:lorentz_geometry}

The $D$-dimensional Lorentz model is the upper sheet of the hyperboloid
\begin{equation}
\mathcal{H}_c^D
=
\left\{
\mathbf{z}\in\mathbb{R}^{D+1}:
\langle\mathbf{z},\mathbf{z}\rangle_{\mathrm{L}}=-c^{-1},
\ z_0>0
\right\},
\label{eq:app_lorentz_manifold}
\end{equation}
with Lorentz inner product
\begin{equation}
\langle\mathbf{x},\mathbf{y}\rangle_{\mathrm{L}}
=
-x_0y_0+\sum_{j=1}^{D}x_jy_j.
\label{eq:app_lorentz_inner_product}
\end{equation}
Its tangent space at $\mathbf{x}\in\mathcal{H}_c^D$ is
\begin{equation}
T_{\mathbf{x}}\mathcal{H}_c^D
=
\left\{
\mathbf{u}\in\mathbb{R}^{D+1}:
\langle\mathbf{x},\mathbf{u}\rangle_{\mathrm{L}}=0
\right\}.
\label{eq:app_lorentz_tangent_space}
\end{equation}
For $\mathbf{u}\in T_{\mathbf{x}}\mathcal{H}_c^D$, define
$\|\mathbf{u}\|_{\mathrm{L}}=
\sqrt{\langle\mathbf{u},\mathbf{u}\rangle_{\mathrm{L}}}$. The exponential
map is
\begin{equation}
\begin{aligned}
\exp_{\mathbf{x}}^{\mathrm{L}}(\mathbf{u})
&=
\cosh\left(\sqrt{c}\|\mathbf{u}\|_{\mathrm{L}}\right)\mathbf{x}
\\
&\quad+
\frac{\sinh\left(\sqrt{c}\|\mathbf{u}\|_{\mathrm{L}}\right)}
{\sqrt{c}\|\mathbf{u}\|_{\mathrm{L}}}\mathbf{u}.
\end{aligned}
\label{eq:app_general_lorentz_expmap}
\end{equation}

At the origin
$\mathbf{o}_{\mathrm{L}}=(c^{-1/2},0,\ldots,0)$, a shared tangent
$\mathbf{v}\in\mathbb{R}^{D}$ is identified with
$(0,\mathbf{v})\in T_{\mathbf{o}_{\mathrm{L}}}\mathcal{H}_c^D$. With
$\varrho(\mathbf{v})=\sqrt{c}\|\mathbf{v}\|$,
Equation~(\ref{eq:app_general_lorentz_expmap}) becomes
\begin{equation}
\begin{aligned}
\exp_{\mathbf{o}_{\mathrm{L}}}^{\mathrm{L}}(\mathbf{v})
&=
\Biggl(
c^{-1/2}\cosh(\varrho(\mathbf{v})),
\\[-2pt]
&\qquad
\frac{\sinh(\varrho(\mathbf{v}))}{\varrho(\mathbf{v})}\mathbf{v}
\Biggr).
\end{aligned}
\label{eq:app_lorentz_origin_expmap}
\end{equation}
The spatial multiplier is defined by its continuous limit, $1$, at the origin.
Direct substitution shows that the image satisfies
Equation~(\ref{eq:app_lorentz_manifold}).

For $\mathbf{x},\mathbf{y}\in\mathcal{H}_c^D$, let
\begin{equation}
\alpha(\mathbf{x},\mathbf{y})
=
-c\langle\mathbf{x},\mathbf{y}\rangle_{\mathrm{L}}
\geq 1.
\label{eq:app_lorentz_alpha}
\end{equation}
The geodesic distance and logarithmic map are
\begin{equation}
d_c^{\mathrm{L}}(\mathbf{x},\mathbf{y})
=
c^{-1/2}\mathrm{arcosh}(\alpha),
\label{eq:app_lorentz_distance}
\end{equation}
\begin{equation}
\log_{\mathbf{x}}^{\mathrm{L}}(\mathbf{y})
=
\frac{\mathrm{arcosh}(\alpha)}{\sqrt{\alpha^2-1}}
(\mathbf{y}-\alpha\mathbf{x}).
\label{eq:app_lorentz_logmap}
\end{equation}
The direction in Equation~(\ref{eq:app_lorentz_logmap}) is tangent at
$\mathbf{x}$ and has squared Lorentz norm $(\alpha^2-1)/c$; consequently,
$\|\log_{\mathbf{x}}^{\mathrm{L}}(\mathbf{y})\|_{\mathrm{L}}
=d_c^{\mathrm{L}}(\mathbf{x},\mathbf{y})$, which verifies the scale factor in
Equation~(\ref{eq:app_lorentz_logmap}).

For $\mathbf{u}\in T_{\mathbf{x}}\mathcal{H}_c^D$, parallel transport along
the unique geodesic from $\mathbf{x}$ to $\mathbf{y}$ is
\begin{equation}
\begin{aligned}
\mathcal{T}_{\mathbf{x}\rightarrow\mathbf{y}}^{\mathrm{L}}(\mathbf{u})
&=
\mathbf{u}
\\
&\quad+
\frac{\langle\mathbf{y},\mathbf{u}\rangle_{\mathrm{L}}}
{c^{-1}-\langle\mathbf{x},\mathbf{y}\rangle_{\mathrm{L}}}
(\mathbf{x}+\mathbf{y}).
\end{aligned}
\label{eq:app_lorentz_parallel_transport}
\end{equation}
The denominator is $(1+\alpha)/c>0$. Parallel transport preserves the Lorentz
norm and maps the vector to $T_{\mathbf{y}}\mathcal{H}_c^D$. HyperANFIS uses
Equation~(\ref{eq:app_lorentz_parallel_transport}) with
$\mathbf{y}=\mathbf{o}_{\mathrm{L}}$ to express all rule-local coordinates in
the shared origin chart.

\subsection{A.3 Poincar\'e Ball and Lorentz Isometry}
\label{app:poincare_geometry}

The Poincar\'e ball of curvature $-c$ is
\begin{equation}
\mathcal{B}_c^D
=
\left\{
\mathbf{q}\in\mathbb{R}^{D}:c\|\mathbf{q}\|^2<1
\right\},
\label{eq:app_poincare_ball}
\end{equation}
with conformal metric
\begin{equation}
g_{\mathbf{q}}^{\mathrm{P}}
=
\lambda_{\mathbf{q}}^2 g^{\mathrm{E}},
\qquad
\lambda_{\mathbf{q}}
=
\frac{2}{1-c\|\mathbf{q}\|^2}.
\label{eq:app_poincare_metric}
\end{equation}
For $\mathbf{z},\mathbf{p}\in\mathcal{B}_c^D$, define
$\Delta_c(\mathbf{q})=1-c\|\mathbf{q}\|^2$. Their distance is
\begin{equation}
\begin{aligned}
d_c^{\mathrm{P}}(\mathbf{z},\mathbf{p})
&=
c^{-1/2}\mathrm{arcosh}
\Biggl(1
\\[-2pt]
&\qquad+
\frac{2c\|\mathbf{z}-\mathbf{p}\|^2}
{\Delta_c(\mathbf{z})\Delta_c(\mathbf{p})}
\Biggr).
\end{aligned}
\label{eq:app_poincare_distance}
\end{equation}

Both hyperbolic models use the same tangent vector at the origin. Under this
convention, the Poincar\'e exponential map is
\begin{equation}
\begin{aligned}
\exp_{\mathbf{o}_{\mathrm{P}}}^{\mathrm{P}}(\mathbf{v})
&=
\frac{\tanh(\varrho(\mathbf{v})/2)}{\varrho(\mathbf{v})}\mathbf{v},
\\
\mathbf{o}_{\mathrm{P}}
&=\mathbf{0}.
\end{aligned}
\label{eq:app_poincare_origin_expmap}
\end{equation}
Its differential at the origin maps $\mathbf{v}$ to $\mathbf{v}/2$. Because
$\lambda_{\mathbf{0}}=2$, the Riemannian norm of this derivative equals the
Euclidean norm of the shared tangent vector $\mathbf{v}$.

The isometry from the Poincar\'e ball to the Lorentz model is
\begin{equation}
\Phi_c(\mathbf{q})
=
\left(
\frac{1+c\|\mathbf{q}\|^2}
{\sqrt{c}(1-c\|\mathbf{q}\|^2)},
\frac{2\mathbf{q}}{1-c\|\mathbf{q}\|^2}
\right),
\label{eq:app_poincare_to_lorentz}
\end{equation}
with inverse
\begin{equation}
\Phi_c^{-1}(\mathbf{z})
=
\frac{\mathbf{z}_{\mathrm{sp}}}{\sqrt{c}z_0+1}.
\label{eq:app_lorentz_to_poincare}
\end{equation}
Using the standard hyperbolic half-angle identities gives
\begin{equation}
\Phi_c\left(
\exp_{\mathbf{o}_{\mathrm{P}}}^{\mathrm{P}}(\mathbf{v})
\right)
=
\exp_{\mathbf{o}_{\mathrm{L}}}^{\mathrm{L}}(\mathbf{v}).
\label{eq:app_expmap_isometry}
\end{equation}
Direct substitution into the Lorentz inner product yields
\begin{equation}
\begin{aligned}
-c\left\langle
\Phi_c(\mathbf{z}),\Phi_c(\mathbf{p})
\right\rangle_{\mathrm{L}}
&=
1
\\[-2pt]
&\qquad+
\frac{2c\|\mathbf{z}-\mathbf{p}\|^2}
{\Delta_c(\mathbf{z})\Delta_c(\mathbf{p})}.
\end{aligned}
\label{eq:app_distance_isometry}
\end{equation}
Equations~(\ref{eq:app_lorentz_distance})--
(\ref{eq:app_distance_isometry}) imply
$d_c^{\mathrm{P}}(\mathbf{z},\mathbf{p})=
d_c^{\mathrm{L}}(\Phi_c(\mathbf{z}),\Phi_c(\mathbf{p}))$. Accordingly,
Poincar\'e distances and tangent-space operations are evaluated after mapping
points through $\Phi_c$. Manifold-valued aggregation results are mapped back
through $\Phi_c^{-1}$. These computations represent the same intrinsic
operations in two coordinate systems.

\section{B Geometry and Membership Variants}
\label{app:model_extensions}

HyperANFIS supports one geometry and one membership function per training run.
The variants share the rule architecture and parameterization; multiple
geometries or membership functions are not combined within a forward pass.

\subsection{B.1 Geometry Variants}
\label{app:geometry_modes}

The Lorentz variant applies Equations~(\ref{eq:app_lorentz_manifold})--
(\ref{eq:app_lorentz_parallel_transport}) directly and is the default. The
Poincar\'e variant applies Equations~(\ref{eq:app_poincare_ball})--
(\ref{eq:app_distance_isometry}); its bounded coordinates represent the same
intrinsic distances as the Lorentz model.

The Euclidean variant preserves the rule architecture but replaces the
manifold operations by
\begin{equation}
\begin{aligned}
\exp_{\mathbf{o}_{\mathrm{E}}}^{\mathrm{E}}(\mathbf{v})&=\mathbf{v},\\
d^{\mathrm{E}}(\mathbf{x},\mathbf{y})&=\|\mathbf{x}-\mathbf{y}\|,\\
\mathbf{e}_{ir}^{\mathrm{E}}&=\mathbf{z}_i^{\mathrm{E}}-\mathbf{p}_r^{\mathrm{E}},\\
\mathbf{m}_i^{\mathrm{E}}&=\sum_{r=1}^{R}\overline{w}_{ir}
\mathbf{q}_{ir}^{\mathrm{E}}.
\end{aligned}
\label{eq:app_euclidean_mode}
\end{equation}
The Euclidean and hyperbolic variants therefore use the same number of rules,
consequent parameterization, and firing normalization. The Euclidean variant
is a geometry-matched counterpart of HyperANFIS, not a conventional
coordinate-wise ANFIS.

\subsection{B.2 Membership Functions}
\label{app:membership_modes}

For any geometry, define the dimension-normalized intrinsic distance as
\begin{equation}
\chi_{ir}
=
\frac{\delta_{ir}^{\mathcal{M}}}{\sqrt{D}\sigma_r}.
\label{eq:app_normalized_rule_distance}
\end{equation}
Each rule scale is bounded through the logit parameterization
\begin{equation}
\sigma_r
=
\sigma_{\min}
+(\sigma_{\max}-\sigma_{\min})\mathrm{sigmoid}(\beta_r).
\label{eq:app_bounded_rule_scale}
\end{equation}
The Gaussian membership function is
\begin{equation}
\ell_{ir}^{\mathrm{G}}
=
-\frac{1}{2}|\chi_{ir}|^2,
\qquad
\mu_{ir}^{\mathrm{G}}=\exp(\ell_{ir}^{\mathrm{G}}).
\label{eq:app_gaussian_log_membership}
\end{equation}
Consequently, the intrinsic radius $\sqrt{D}\sigma_r$ corresponds to
$\mu_{ir}^{\mathrm{G}}=\exp(-1/2)$.

The generalized Bell membership function with fixed $b>0$ is
\begin{equation}
\ell_{ir}^{\mathrm{B}}
=
-\log\left(1+|\chi_{ir}|^{2b}\right),
\qquad
\mu_{ir}^{\mathrm{B}}
=
\frac{1}{1+|\chi_{ir}|^{2b}}.
\label{eq:app_bell_log_membership}
\end{equation}
Both functions equal $1$ at the rule center. The Gaussian tail decays
exponentially in $\chi_{ir}^2$, whereas the Bell tail decays polynomially as
$|\chi_{ir}|^{-2b}$. The Bell shape parameter $b$ controls the transition
between the central region and the tail; the default is $b=2$.

Normalized firing strengths are computed in the log domain:
\begin{equation}
\begin{aligned}
\overline{w}_{ir}
&=
\frac{\exp(\ell_{ir})}
{\sum_{q=1}^{R}\exp(\ell_{iq})}\\
&=
\exp\left(
\ell_{ir}-\mathrm{logsumexp}_{q}(\ell_{iq})
\right).
\end{aligned}
\label{eq:app_log_domain_firing}
\end{equation}
This form does not require explicit exponentiation before normalization and
therefore avoids underflow of intermediate membership values. Numerical
safeguards are specified in Appendix~\ref{app:numerical_safeguards}.

\section{C Inference and Optimization}
\label{app:complete_optimization}

This section specifies rule-local consequent construction, intrinsic
aggregation, the training objective, and gradient computation.

\subsection{C.1 Rule-Local Consequents}
\label{app:local_consequents}

For $\mathcal{M}\in\{\mathrm{L},\mathrm{P}\}$, the local coordinate of sample
$i$ relative to rule $r$ is
\begin{equation}
\begin{aligned}
\boldsymbol{\xi}_{ir}
&=
\log_{\mathbf{p}_r^{\mathcal{M}}}^{\mathcal{M}}
(\mathbf{z}_i^{\mathcal{M}}),\\
\boldsymbol{\eta}_{ir}
&=
\mathcal{T}_{\mathbf{p}_r^{\mathcal{M}}\rightarrow
\mathbf{o}_{\mathcal{M}}}^{\mathcal{M}}
\left(\boldsymbol{\xi}_{ir}\right),\\
\mathbf{e}_{ir}
&=
\left[\boldsymbol{\eta}_{ir}\right]_{\mathrm{sp}}.
\end{aligned}
\label{eq:app_local_rule_coordinate}
\end{equation}
For the Lorentz model, the logarithmic map and parallel transport are given by
Equations~(\ref{eq:app_lorentz_logmap}) and
(\ref{eq:app_lorentz_parallel_transport}). For the Poincar\'e model, both
points are mapped by $\Phi_c$, the Lorentz operations are applied, and the
shared spatial tangent coordinates are retained. For the Euclidean model,
$\mathbf{e}_{ir}=\mathbf{z}_i-\mathbf{p}_r$.

The first-order consequent tangent and its manifold image are
\begin{equation}
\begin{aligned}
\mathbf{u}_{ir}
&=
\mathrm{clip}_{\tau}
\left(\mathbf{b}_r+\mathbf{W}_r\mathbf{e}_{ir}\right),\\
\mathbf{q}_{ir}^{\mathcal{M}}
&=
\exp_{\mathbf{o}_{\mathcal{M}}}^{\mathcal{M}}(\mathbf{u}_{ir}),
\end{aligned}
\label{eq:app_rule_consequent}
\end{equation}
where $\mathbf{b}_r\in\mathbb{R}^{H}$ and
$\mathbf{W}_r\in\mathbb{R}^{H\times D}$. Setting
$\mathbf{W}_r=\mathbf{0}$ gives the zero-order variant. Each consequent remains
rule-specific through its base point $\mathbf{p}_r$ and affine parameters
$(\mathbf{b}_r,\mathbf{W}_r)$, although aggregation uses a shared origin chart.

\subsection{C.2 Fr\'echet Aggregation and Classification}
\label{app:frechet_details}

The aggregate for sample $i$ minimizes the weighted Fr\'echet objective
\begin{equation}
\begin{aligned}
\mathcal{F}_i(\mathbf{m})
&=
\frac{1}{2}\sum_{r=1}^{R}\overline{w}_{ir}
d_c^{\mathcal{M}}
(\mathbf{m},\mathbf{q}_{ir}^{\mathcal{M}})^2,\\
\mathbf{m}_i^{\star}
&=\mathop{\mathrm{argmin}}_{\mathbf{m}\in\mathcal{M}}
\mathcal{F}_i(\mathbf{m}).
\end{aligned}
\label{eq:app_frechet_objective}
\end{equation}
The Riemannian gradient is
$\mathrm{grad}\,\mathcal{F}_i(\mathbf{m})=
-\sum_r\overline{w}_{ir}\log_{\mathbf{m}}(\mathbf{q}_{ir})$.
The update below is therefore Riemannian gradient descent with step size
$\eta_{\mathrm{K}}$. The Fr\'echet minimizer is unique in hyperbolic space, but
the finite Karcher iterations yield a differentiable approximation to the
minimizer.

For Lorentz consequents, the normalized ambient initialization is
\begin{equation}
\begin{aligned}
\mathbf{h}_i
&=
\sum_{r=1}^{R}\overline{w}_{ir}\mathbf{q}_{ir}^{\mathrm{L}},\\
\mathbf{m}_i^{(0)}
&=
\frac{\mathbf{h}_i}
{\sqrt{c\left(-\langle\mathbf{h}_i,\mathbf{h}_i\rangle_{\mathrm{L}}\right)}}.
\end{aligned}
\label{eq:app_barycenter_initialization}
\end{equation}
This normalization maps $\mathbf{h}_i$ to $\mathcal{H}_c^H$. Because the positive
firing weights sum to one, $\mathbf{h}_i$ remains in the future-directed
timelike cone and the normalization selects the upper sheet. Poincar\'e
consequents are first mapped through
$\Phi_c$, initialized by Equation~(\ref{eq:app_barycenter_initialization}), and
mapped back after refinement. The Euclidean model uses the weighted arithmetic
mean in Equation~(\ref{eq:app_euclidean_mode}).

Starting from $\mathbf{m}_i^{(0)}$, HyperANFIS computes
\begin{equation}
\begin{aligned}
\mathbf{d}_i^{(k)}
&=
\sum_{r=1}^{R}\overline{w}_{ir}
\log_{\mathbf{m}_i^{(k)}}^{\mathcal{M}}
(\mathbf{q}_{ir}^{\mathcal{M}}),\\
\mathbf{m}_i^{(k+1)}
&=
\exp_{\mathbf{m}_i^{(k)}}^{\mathcal{M}}
(\eta_{\mathrm{K}}\mathbf{d}_i^{(k)}).
\end{aligned}
\label{eq:app_karcher_update}
\end{equation}
The iterations terminate when $\|\mathbf{d}_i^{(k)}\|\leq
\epsilon_{\mathrm{K}}$ for every sample in the batch or after three steps.
At a stationary point,
$\sum_r\overline{w}_{ir}\log_{\mathbf{m}_i}(\mathbf{q}_{ir})=\mathbf{0}$,
which is the first-order condition of Equation~(\ref{eq:app_frechet_objective}).

For classification, class tangents $\mathbf{g}_k\in\mathbb{R}^{H}$ define
\begin{equation}
\begin{aligned}
\mathbf{c}_k^{\mathcal{M}}
&=
\exp_{\mathbf{o}_{\mathcal{M}}}^{\mathcal{M}}
\left[\mathrm{clip}_{\tau}(\mathbf{g}_k)\right],\\
s_{ik}
&=
-d_c^{\mathcal{M}}
(\mathbf{m}_i^{\mathcal{M}},\mathbf{c}_k^{\mathcal{M}})^2.
\end{aligned}
\label{eq:app_classification_logits}
\end{equation}
\subsection{C.3 Training Objective}
\label{app:loss_definitions}

For a mini-batch of size $B$, let
$P_{ik}=\mathrm{softmax}_{k}(s_{ik})$ and let $y_i$ be the class label. If
$n_k$ is the number of training samples in class $k$, the class weight is
\begin{equation}
\gamma_k
=
\frac{N}{K n_k},
\label{eq:app_class_weight}
\end{equation}
where $N$ and $K$ are the number of training samples and classes, respectively.
The classification loss is
\begin{equation}
\mathcal{L}_{\mathrm{CE}}
=
-\frac{1}{B}
\sum_{i=1}^{B}\gamma_{y_i}\log P_{i y_i}.
\label{eq:app_weighted_cross_entropy}
\end{equation}
Class weighting is applied only during training. Validation loss and macro-F1
are unweighted.

Define the batch-average rule usage
\begin{equation}
\pi_r
=
\frac{1}{B}\sum_{i=1}^{B}\overline{w}_{ir}.
\label{eq:app_rule_usage}
\end{equation}
The rule-balance term is the divergence of average rule usage from the uniform
distribution:
\begin{equation}
\mathcal{L}_{\mathrm{bal}}
=
\sum_{r=1}^{R}\pi_r
\log\left(\frac{\pi_r}{1/R}\right).
\label{eq:app_balance_loss}
\end{equation}
This term is minimized by uniform average usage and does not require uniform
firing for each sample.

The rule-specialization term is the normalized firing entropy:
\begin{equation}
\mathcal{L}_{\mathrm{spec}}
=
-\frac{1}{B\log R}
\sum_{i=1}^{B}\sum_{r=1}^{R}
\overline{w}_{ir}\log\overline{w}_{ir}.
\label{eq:app_specialization_loss}
\end{equation}
Minimizing this term encourages concentrated sample-level firing. Its
coefficient follows the warm-up schedule
\begin{equation}
\kappa(t)
=
\min\left(\frac{t}{T_{\mathrm{w}}},1\right),
\label{eq:app_specialization_warmup}
\end{equation}
which delays the full specialization penalty while rule prototypes establish
coverage. The balance term discourages globally unused rules, whereas the
specialization term discourages diffuse sample-level firing.

The prototype-separation term is
\begin{equation}
\begin{aligned}
\mathcal{L}_{\mathrm{sep}}
&=
\frac{1}{R(R-1)}
\sum_{r\neq q}
\\[-2pt]
&\qquad\cdot
\left[
m_0-d_c^{\mathcal{M}}
(\mathbf{p}_r^{\mathcal{M}},\mathbf{p}_q^{\mathcal{M}})
\right]_{+}^{2}.
\end{aligned}
\label{eq:app_separation_loss}
\end{equation}
Only prototype pairs separated by less than $m_0$ contribute. The training
objective is
\begin{equation}
\begin{aligned}
\mathcal{L}(t)
&=
\mathcal{L}_{\mathrm{CE}}
+\lambda_{\mathrm{bal}}\mathcal{L}_{\mathrm{bal}}
\\
&\quad
+\kappa(t)\lambda_{\mathrm{spec}}\mathcal{L}_{\mathrm{spec}}
+\lambda_{\mathrm{sep}}\mathcal{L}_{\mathrm{sep}}.
\end{aligned}
\label{eq:app_complete_training_loss}
\end{equation}

\subsection{C.4 Optimization}
\label{app:backpropagation}

The forward map is differentiable almost everywhere with respect to the
tangent-space centers, scale logits, consequent parameters, and class
prototypes. For either membership function, firing normalization satisfies
\begin{equation}
\frac{\partial\overline{w}_{ir}}
{\partial\ell_{iq}}
=
\overline{w}_{ir}
\left(\mathbf{1}[r=q]-\overline{w}_{iq}\right).
\label{eq:app_softmax_derivative}
\end{equation}
The log-membership derivatives are
\begin{equation}
\begin{aligned}
\frac{\partial\ell^{\mathrm{G}}}{\partial\chi}
&=-\chi,\\
\frac{\partial\ell^{\mathrm{B}}}{\partial\chi}
&=
-\frac{2b\,\mathrm{sign}(\chi)|\chi|^{2b-1}}
{1+|\chi|^{2b}}.
\end{aligned}
\label{eq:app_membership_derivatives}
\end{equation}
The sigmoid parameterization in Equation~(\ref{eq:app_bounded_rule_scale})
keeps each scale trainable within its prescribed interval. Radial clipping is
the identity inside the tangent ball and removes the outward radial gradient
component outside the ball. Its nondifferentiable boundary has measure zero;
automatic differentiation uses the active branch.

The Karcher iterations in Equation~(\ref{eq:app_karcher_update}) are unrolled
in the computation graph. Gradients propagate through the logarithmic maps,
weighted tangent sums, exponential updates, and ambient initialization. The
discrete nearest-rule assignments are computed before training and are not
differentiated. Early stopping and collapse detection do not modify the forward
map.

The numerical clamps in Appendix~\ref{app:numerical_safeguards} are piecewise
differentiable. An active clamp has zero gradient toward the invalid region,
preventing updates across the Poincar\'e boundary or the domain boundary of
$\mathrm{arcosh}$.

Preprocessing and nearest-rule initialization precede gradient-based training.
For each mini-batch, the model computes manifold embeddings, memberships,
firing strengths, rule consequents, and the Karcher aggregate; evaluates
Equation~(\ref{eq:app_complete_training_loss}); and applies one optimizer update
after backpropagation. Model-selection criteria are given in
Appendix~\ref{app:training_control}.

\section{D Theoretical Analysis}
\label{app:theoretical_properties}

We establish the uniqueness of intrinsic aggregation and characterize the
zero-curvature limit of the hyperbolic distance.

\subsection{D.1 Uniqueness of Fr\'echet Aggregation}
\label{app:aggregation_properties}

\paragraph{Proposition 1.}
For nonnegative weights $\overline{w}_{ir}$ summing to one, the Fr\'echet
objective in Equation~(\ref{eq:app_frechet_objective}) has a unique minimizer
in the Lorentz and Poincar\'e models.

\paragraph{Proof.}
Hyperbolic space is complete, simply connected, and has constant negative
sectional curvature and is therefore a Hadamard manifold. On a Hadamard
manifold, the weighted sum of squared distances is coercive and strictly
geodesically convex. Completeness and coercivity give existence, and strict
convexity gives uniqueness. The result holds in both coordinate models by the
isometry $\Phi_c$. In Euclidean space, the corresponding objective is a convex
quadratic with minimizer
$\sum_r\overline{w}_{ir}\mathbf{q}_{ir}$. \hfill$\square$

HyperANFIS approximates this minimizer with at most three Karcher iterations and
terminates earlier when the batch-wide tolerance is satisfied.

\subsection{D.2 Euclidean Limit}
\label{app:euclidean_limit}

\paragraph{Proposition 2.}
For fixed shared tangent vectors $\mathbf{v}$ and $\mathbf{a}$,
\begin{equation}
\lim_{c\downarrow0}
d_c^{\mathrm{L}}
\Biggl(
\begin{aligned}
&
\exp_{\mathbf{o}_{\mathrm{L}}}^{\mathrm{L}}(\mathbf{v}),
\\[-2pt]
&
\exp_{\mathbf{o}_{\mathrm{L}}}^{\mathrm{L}}(\mathbf{a})
\end{aligned}
\Biggr)
=\|\mathbf{v}-\mathbf{a}\|.
\label{eq:app_zero_curvature_limit}
\end{equation}

\paragraph{Proof.}
Expanding the hyperbolic sine and cosine terms in
Equation~(\ref{eq:app_lorentz_origin_expmap}) gives
\begin{align}
&-c\left\langle
\begin{aligned}
&
\exp_{\mathbf{o}_{\mathrm{L}}}^{\mathrm{L}}(\mathbf{v}),
\\[-2pt]
&
\exp_{\mathbf{o}_{\mathrm{L}}}^{\mathrm{L}}(\mathbf{a})
\end{aligned}
\right\rangle_{\mathrm{L}}
\nonumber\\
&\qquad=
1+\frac{c}{2}\|\mathbf{v}-\mathbf{a}\|^2+O(c^2).
\label{eq:app_lorentz_inner_expansion}
\end{align}
Since $\mathrm{arcosh}(1+u)=\sqrt{2u}+O(u^{3/2})$ as $u\downarrow0$,
substitution into Equation~(\ref{eq:app_lorentz_distance}) proves the claim.
The same limit holds in Poincar\'e coordinates by isometry.
\hfill$\square$

Under the tangent bounds in Equation~(\ref{eq:app_tangent_clipping}), the
logarithmic maps, transported local coordinates, and Fr\'echet objective also
converge to their Euclidean counterparts.
The Euclidean variant is therefore the zero-curvature counterpart of the global
radial-rule architecture, not a conventional ANFIS based on products of
coordinate-wise memberships.

\section{E Implementation Details}
\label{app:implementation_details}

This section specifies preprocessing, initialization, numerical safeguards,
and model selection.

\subsection{E.1 Preprocessing and Initialization}
\label{app:input_scaling_and_initialization}

All preprocessing statistics are estimated from the training partition. For raw
feature $x_{ij}$,
\begin{equation}
\widehat{x}_{ij}
=
\frac{x_{ij}-\nu_j}{\zeta_j},
\label{eq:app_standardization}
\end{equation}
where $\nu_j$ and $\zeta_j$ are the training mean and standard deviation. Let
$Q_{0.95}$ denote the empirical $0.95$ quantile and let $\rho$ be the target
tangent radius. The fixed input scale is
\begin{equation}
\begin{aligned}
\mathcal{R}_{\mathrm{tr}}
&=
\left\{\|\widehat{\mathbf{x}}_i\|:
i\in\mathcal{D}_{\mathrm{tr}}\right\},\\
r_{0.95}
&=
\max\left\{
Q_{0.95}(\mathcal{R}_{\mathrm{tr}}),10^{-8}
\right\},\\
s
&=\frac{\rho}{r_{0.95}}.
\end{aligned}
\label{eq:app_automatic_input_scale}
\end{equation}
This normalization maps the empirical reference radius to $\rho$ before radial
clipping; $\tau$ remains a separate upper bound.

Initial center tangents are sampled from
$\{\mathrm{clip}_{\tau}(s\widehat{\mathbf{x}}_i)\}$ without replacement when
the training set contains at least $R$ samples, and with replacement otherwise.
For at most $N_{\mathrm{init}}$ randomly selected training samples, define
\begin{equation}
\begin{aligned}
r^{\star}(i)
&=
\mathop{\mathrm{argmin}}_{1\leq r\leq R}\delta_{ir}^{\mathcal{M}},
\\
d_i^{\mathrm{near}}
&=
\frac{1}{\sqrt{D}}
\min_{1\leq r\leq R}\delta_{ir}^{\mathcal{M}}.
\end{aligned}
\label{eq:app_nearest_rule_assignment}
\end{equation}
The global scale estimate is the $q_{\mathrm{init}}$ quantile of
$\{d_i^{\mathrm{near}}\}$. Rule $r$ uses the same quantile of
$\{\delta_{ir}^{\mathcal{M}}/\sqrt{D}:r^{\star}(i)=r\}$. Empty or invalid per-rule estimates
are replaced by a positive global estimate. After multiplication by
$m_{\mathrm{init}}$, scales are clipped to
$[\sigma_{\min},\sigma_{\max}]$ and converted to the logit parameterization in
Equation~(\ref{eq:app_bounded_rule_scale}).

Consequent biases and class-prototype tangents are initialized from
$\mathcal{N}(0,0.05^2)$ and $\mathcal{N}(0,0.15^2)$, respectively. Consequent
matrices use Xavier-uniform initialization with gain $0.25$. These initialized
quantities remain trainable, whereas sample-to-rule assignments are used only
during initialization.

\subsection{E.2 Numerical Stability}
\label{app:numerical_safeguards}

Poincar\'e points are radially capped at
$(1-\epsilon_{\mathrm{B}})/\sqrt{c}$ with
$\epsilon_{\mathrm{B}}=10^{-5}$. Lorentz projection preserves the spatial
coordinates and recomputes the positive time coordinate as
$\sqrt{c^{-1}+\|\mathbf{z}_{\mathrm{sp}}\|^2}$. The radial norm in the
Poincar\'e projection uses the lower bound
$\epsilon_{\mathrm{N}}=10^{-12}$.

Denominators, squared norms, and nonnegative radicands use the lower bound
$\epsilon_{\mathrm{N}}$. Arguments to $\mathrm{arcosh}$ use
$1+\epsilon_{\mathrm{A}}$, where $\epsilon_{\mathrm{A}}=10^{-7}$. This lower
bound is an implementation safeguard; analytically,
$\mathrm{arcosh}(1)=0$. Explicit membership
values use a lower bound of $10^{-12}$ for reporting, whereas firing strengths
are computed from the unclamped log-memberships in
Equation~(\ref{eq:app_log_domain_firing}). Karcher iterates are reprojected
after each step, and Poincar\'e aggregation is evaluated through the Lorentz
isometry.

\subsection{E.3 Feature-Space Interpretation}
\label{app:raw_replay}

For a raw observation $\mathbf{x}$, antecedent evaluation follows
\begin{equation}
\begin{aligned}
\mathbf{x}
&\rightarrow 
\widehat{\mathbf{x}}
=
(\mathbf{x}-\boldsymbol{\nu})\oslash\boldsymbol{\zeta}\\
&\rightarrow 
\mathbf{v}
=
\mathrm{clip}_{\tau}(s\widehat{\mathbf{x}})\\
&\rightarrow 
\mathbf{z}^{\mathcal{M}}
=
\exp_{\mathbf{o}_{\mathcal{M}}}^{\mathcal{M}}(\mathbf{v}),
\end{aligned}
\label{eq:app_raw_feature_replay}
\end{equation}
where $\oslash$ denotes element-wise division. Reproducing this mapping requires
the fitted preprocessing statistics and the scale $s$.

A rule center has shared tangent representative
$\overline{\mathbf{a}}_r=\mathrm{clip}_{\tau}(\mathbf{a}_r)$. Its
feature-space representative is
\begin{equation}
\widetilde{\mathbf{x}}_r
=
\boldsymbol{\nu}
+\boldsymbol{\zeta}\odot
\frac{\overline{\mathbf{a}}_r}{s},
\label{eq:app_rule_center_raw_representative}
\end{equation}
where $\odot$ denotes element-wise multiplication. This inverse gives a
feature-space representative of the rule center. It is not globally unique,
because inputs outside the tangent bound can map to the same clipped point.

\subsection{E.4 Model Selection}
\label{app:training_control}

Validation macro-F1 controls learning-rate scheduling, checkpointing, and early
stopping. The test set is evaluated once with the checkpoint that attains the
highest validation macro-F1. After a fixed grace period, a diagnostic terminates
persistent single-class collapse without modifying
Equation~(\ref{eq:app_complete_training_loss}).

\end{document}